\documentclass{article}
\usepackage{spconf,amsmath,graphicx, animate}
\usepackage{tabularx}
\usepackage{hyperref}
\usepackage{url}
\usepackage{pdfpages}

\newcommand{\etal}{\textit{et al.}}

\hypersetup{colorlinks=true,citecolor=green}

\title{Saliency Tubes: Visual Explanations for Spatio-Temporal Convolutions}
\name{\normalsize Alexandros Stergiou$^{1 \dagger}$\thanks{$^{\dagger}$Corresponding author}, Georgios Kapidis$^{1,4}$, Grigorios Kalliatakis$^{2}$, Christos Chrysoulas $^{3}$, Remco Veltkamp$^{1}$, Ronald Poppe$^{1}$}

\address{
$^{1}$ Department of Information and Computing Sciences, Utrecht University, Utrecht, Netherlands\\
$^{2}$ School of Computer Science and Electronic Engineering, University of Essex, Colchester, United Kingdom\\
$^{3}$ School of Computer Science and Informatics, London South Bank University, London, United Kingdom\\
$^{4}$ Noldus Information Technology, Wageningen, The Netherlands \\
\normalsize \{a.g.stergiou, g.kapidis, r.c.veltkamp,  r.w.poppe\}@uu.nl, gkallia@essex.ac.uk, chrysouc@lsbu.ac.uk}
\begin{document}
%
\maketitle
\begin{abstract}
Deep learning approaches have been established as the main methodology for video classification and recognition. Recently, 3-dimensional convolutions have been used to achieve state-of-the-art performance in many challenging video datasets. Because of the high level of complexity of these methods, as the convolution operations are also extended to an additional dimension in order to extract features from it as well, providing a visualization for the signals that the network interpret as informative, is a challenging task. An effective notion of understanding the network's inner-workings would be to isolate the spatio-temporal regions on the video that the network finds most informative. We propose a method called \textit{Saliency Tubes} which demonstrate the foremost points and regions in both frame level and over time that are found to be the main focus points of the network. We demonstrate our findings on widely used datasets for third-person and egocentric action classification and enhance the set of methods and visualizations that improve 3D Convolutional Neural Networks (CNNs) intelligibility. Our code \footnote{\url{https://goo.gl/xX4nnv}} and a demo video \footnote{\url{https://youtu.be/JANUqoMc3es}} are also available. 


\end{abstract}
\section{Introduction}
\label{sec:intro}

Deep Convolutional Neural Networks  (CNNs) have enabled unparalleled breakthroughs in a variety of visual tasks, such as  image classification \cite{he2016deep,krizhevsky2012imagenet}, object detection \cite{girshick2014rich}, image captioning \cite{chen2015microsoft,vinyals2015show}, and video classification \cite{simonyan2014two,gkioxari2015finding, stergiou2018understanding}. While these deep neural networks show superior performance, they are often criticized as black boxes that lack interpretability, because of their end-to-end learning approach. This hinders the understanding of which features are extracted and what improvements can be made in the architectural level.

Hence, there has been a significant interest over the last few years in developing various methods of interpreting CNN models \cite{chattopadhay2018grad,montavon2018methods,selvaraju2017grad,ribeiro2016should}. One such category of methods probes the neural network models by trying to change the input and analyzing the model's response to it. Another approach is to explain the decision of a model by training another deep model which reveals the visual explanations. 

While there has been promising progress in the context of these 'visual explanations' for 2D CNNs, visualizing learned features of 3D convolutions, where the networks have access to not only the appearance information present in single, static images, but also their complex temporal evolution, has not received the same attention. To extend 'visual explanations' to spatio-temporal data such as videos, we propose Saliency Tubes, a generalized attention mechanism for explaining CNN decisions, which is inspired by the class activation mapping (CAM) proposed in \cite{zhou2016learning}. 

Saliency Tubes is a general and extensible module that can be easily plugged into any existing spatio-temporal CNN architecture to enable human-interpretable visual explanations across multiple tasks including action classification and egocentric action recognition.

\begin{figure*}[!h]
\centering
\includegraphics[width=.8 \textwidth]{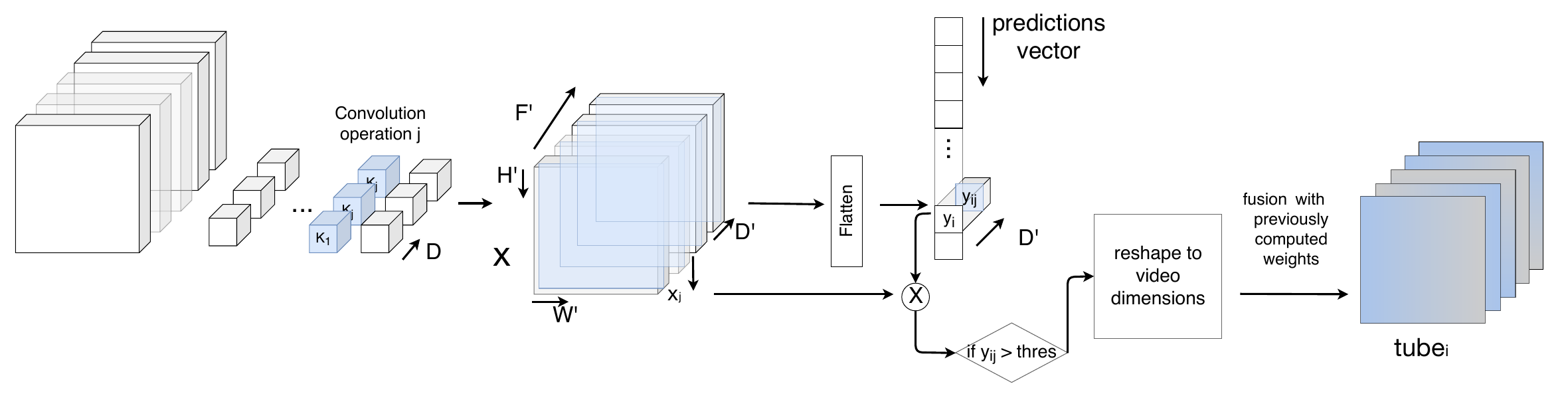}
\caption{\textbf{Saliency Tubes.} (a) Informative regions are found based on the activation maps from output $x$, while the useful features are defined based on their corresponding values in feature vector $y_{i}$. (b) Individual features can also be visualised spatio-temporally by fusing all the activation maps in the last step of the focus tubes and only include single re-scaled activation maps. The illustration is presented for the simplified case of a single convolution layer for convenience and easier interpretability. }
\label{fig:Fig1}
\end{figure*}

Our key contributions are summarized as follows:
\begin{itemize}
    \item We propose Saliency Tubes, a spatio-temporal-specific, class-discriminative technique that generates visual explanations from any 3D ConvNet without requiring architectural changes or re-training. 
    \item We apply Saliency Tubes to existing top-performing video recognition spatio-temporal models. For action classification, our visualizations highlight the important regions in the video for predicting the action class, and shed light on why the predictions succeed or fail. For egocentric action recognition, our visualizations point out the target objects of the overall motion, and how their interaction with the position of the hands indicates patterns of everyday actions.
    \item Through visual examples, we show that Saliency Tubes improve upon the region-specific nature of Map methods by showing a generalized spatio-temporal focus of the network.
\end{itemize}

Related work on visual interpretability for neural network representations is summarized in Section \ref{sec:relatedwork}. In Section \ref{sec:focustubes} the details of the proposed approach are presented. In Section \ref{sec:resuls}, we  report the visualization results in third person and egocentric action classification and we discuss their descriptiveness. The paper's main conclusions are drawn in Section \ref{sec:conclusions}.

\section{Related Work}
\label{sec:relatedwork}

Bau \etal~argue \cite {bau2017network}, that two of the key elements of CNNs should be their discrimination capabilities and their interpretability. Although the discrimination capabilities of CNNs have been well established, the same could not be said about their interpretability as ways of visualizing each of their binding parts have been proven challenging. The direct visualization of convolutional kernels has been a well explored field with many works based  on the inversion of feature maps to images \cite{dosovitskiy2016inverting} as well as gradient-based visualizations \cite{zeiler2014visualizing, mahendran2015understanding}. Following the gradient-based approaches, one of the first attempts to present the network's receptive field was proposed by Zhou \etal~\cite{zhou2014object} in which the output neural activation feature maps were represented with image-resolution. Others have also focused on parts of networks and how specific inputs can be used to identify the units that have larger activations \cite{yosinski2015understanding}. Konam \cite{konam2017vision} proposed a method for discovering regions that excite particular neurons. This notion was later used as the main point for creating explanatory graphs, which correspond to features that are tracked though the CNN's extracted feature hierarchy \cite{zhang2018interpreting, zhang2017growing} and are based on separating the different feature parts of convolutional kernels and representing individually the parts of different extracted kernel patterns.

Only few works have addressed the video domain, aimed at reproducing the visual explanations achieved in image-based models. Karpathy \etal~\cite{karpathy2015visualizing} visualized the Long Short Term Memory (LSTM) cells. Bargal \etal~\cite{adel2018excitation} have studied class activations for action recognition in systems composed of 2D CNN classifiers combined with LSTMs for monitoring the temporal variations of the CNN outputs. Their approach was based on Excitation Backpropagation \cite{zhang2018top}. Their main focus was based on the decision made by  Recurrent Neural Networks (RNNs) in action recognition and video captioning, with the convolutional blocks only being used as per-frame feature extractors. In 3D action recognition, Chattopadhay~\etal \cite{chattopadhay2018grad} have proposed a generalized version of the class activation maps generalized for object recognition.

To address the lack of visual explanation methods for 3D convolutions, we propose Saliency Tubes, which are constructed for finding both regions and frames the network focuses on. This method can be generalized to different action-based approaches in videos as we demonstrate for both third person and egocentric tasks.

\section{Saliency Tubes}
\label{sec:focustubes}

Figure \ref{fig:Fig1} outlines our approach. We let $x$ denote the activation maps of the network's final convolutional layer with output maps of size $F' \times W' \times H' \times D'$, where $F'$ represents the number of frames that are used, $W'$ is the width of the activation maps, $H'$ is the height and $D'$ is the number of channels (also could be referred to as frame-wide depth) that equal the total number of convolutional operations performed in that layer. Let also $y_{i}$ be the tensor in the final fully-connected layer responsible for class predictions, of a specific class $i$ (with $i \in \{0,N\}$ and $N$ being the total number of classes). We consider every element of the predictions vector, denoted as $y_{i,j}$ which corresponds to a specific depth dimension of the network's final convolutional layer ($x$) and designates how informative that specific activation map is towards a correct prediction for an example of class $i$. In order to do so, we propagate back to these activation maps ($a_{f, w, h, j}$) and multiply all their elements by the equivalent predictions weight vector $y_{i,j}$. The class weighted operation can be formulated as $z_{i,j}$ in which:

\begin{equation}
\label{eq:eq1}
\begin{split}
z_{i,j} = \sum_{f}^{F'} \sum_{w}^{W'} \sum_{h}^{H'} y_{i,j} \times a_{f, w, h, j} \quad \forall \:
y_{i,j} \geq \tau
\end{split}
\end{equation}

Because of the large number of features that are extracted by the network (dimension $D'$ could take a value in the range of thousands in modern architectures), we specify a threshold $\tau$ based on which, only the activations that significantly contribute to the predictions are selected. We define all values below this threshold as elements of set $E$.

Following the matrix multiplication to find the feature's intensity, the activations are then reshaped to correspond to the original video dimensions of $F \times W \times H$. During the reshape process we use spline interpolation for increasing the spatio-temporal dimensions of the thresholded activations. To create the final saliency tubes, the operation described in Equation~\ref{eq:eq1} is performed for $j$ features with the final output being:

\begin{equation}
\label{eq:eq2}
\begin{split}
tube_{i} = \sum_{j}^{D'} z_{i,j}, \quad \forall \: j \in \{0,D'\} \notin E. 
\end{split}
\end{equation}

\section{Visualization of Saliency Tubes}
\label{sec:resuls}

In Section~\ref{sec:resuls_localization} we visualize the outputs of 3D CNNs using Saliency Tubes in two different forms and in Section ~\ref{sec:resuls_cnn_compar} we compare them against the outputs of 2D CNNs.      

\subsection{Localization of Saliency Tubes}
\label{sec:resuls_localization}

In Figure~\ref{fig:Fig2}, we demonstrate two cases of video activity classification with overlaid Heat and Focus Tubes, which we utilize as a means to visualize the Saliency Tubes. To produce the activation maps we use a 3D Multi-Fiber Network (MFNet) \cite{chen2018multi} pretrained on Kinetics \cite{carreira2017quo} and subsequently finetuned on UCF-101 \cite{soomro2012ucf101} and  EPIC-Kitchens (verbs) \cite{damen2018scaling}, respectively. Our aim is to examine the regions in space and time that the networks focus on for a particular action class and feature. In our examples, the network input is 16 frames, which we also use as a visualization basis to overlay the output.

In row 1 of Figure~\ref{fig:Fig2}, we show the example of a person performing a martial arts exhibition (TaiChi class), from the test set of UCF-101 \cite{soomro2012ucf101}. The Saliency Tubes show that the network does not fixate on the person but follows parts that correlate with the movement as it progresses. We observe high activations during the backstep and left-hand motions, but not the front-step in between. This shows that the network finds some specific action segments more informative than others, in terms of selected classes instead of the whole range of motions that exist in the video.

\begin{figure}[htb]
    \hskip 1pt Original video \hskip 20pt Heat Tube \hskip 27pt Focus Tube \\
    \centering
    \animategraphics[loop,autoplay,height=60pt]{10}{join_videos_and_ego/3rdperson/combo_comp/}{0}{15} \\
    \animategraphics[loop,autoplay,height=60pt]{10}{join_videos_and_ego/ego/combo_comp/}{0}{15} 
\caption{\textbf{Visualizing Saliency Tubes}. Row 1 presents examples from 3rd person perspective videos, such as those found in UCF-101 \cite{soomro2012ucf101}. For the second row we focus on egocentric tasks from the EPIC-Kitchens dataset \cite{damen2018scaling}. For both examples, we use a 3D Multi-Fiber Network \cite{chen2018multi}, pre-trained on the Kinetics dataset \cite{carreira2017quo} and finetuned on each of the two datasets. Viewed better on Adobe Reader where the subfigures play as videos.}
\label{fig:Fig2}
\end{figure}

In Fig.~\ref{fig:Fig2} row 2, we visualize a segment from the EPIC-Kitchens \cite{damen2018scaling} dataset with the action label 'open door'. Here, our classifier is trained only on verb classes, therefore we expect it to consider motions as more significant than appearance features when predicting a class label. Initially, the moving hand is shown to produce relatively high activations; significantly higher compared to the 'door' area which is the main object of the segment. After a period of movement towards the door that is not considered meaningful, high activations are correlated with the door's movement. This leads to the realization that the network notices this movement and takes these features into account for the class prediction. It is important to note that the focus of the activations does not depend solely on the moving object, but is largely dependant on the area of the motion. Finally, as the door moves out of the scene, the activations remain high in the area in which it used to be. This analysis of a 3D network's output is only possible due to Saliency Tube's ability to visualize its activations as a whole and not per frame.

\subsection{Saliency comparison of 2D and 3D Convolutions}
\label{sec:resuls_cnn_compar}

We further compare our results to the ones obtained by directly using 2D convolutions. More specifically, we use a Temporal Segment Network (TSN) \cite{wang2016temporal} pre-trained on ImageNet \cite{krizhevsky2012imagenet} and finetuned on EPIC-Kitchens (verbs) \cite{damen2018scaling} to demonstrate the class activations in videos of 2D convolutions, while we use the MFNet \cite{chen2018multi} from our previous example for spatio-temporal activations. In Figure~\ref{fig:Fig3}, a bounding box annotation from \cite{damen2018scaling} is regarded as the possible area of interest in the scene which we overlay on the corresponding heat-maps \cite{selvaraju2017grad} and heat-tubes of the final convolutional layer from each network respectively. The heat-maps created were based on slight modifications in our method in order to correspond with the decreased tensor dimensionality.

The 2D convolutions from TSN show time-invariant activations, meaning that the model will make class predictions based on appearance features in every frame. Therefore, the movement occurring in the action is not taken into account, making the predictions to depend heavily on both model complexity (as for overfitting) and strong inter-class similarities. This also empowers the notion of using supplementary crafted temporal features (such as optical flow) for including motion features as input for the network. In contrast, Saliency Tubes exhibit that temporal movement is highly influential to 3D convolutions when determining class features. Our visualizations confirm that alongside finding regions in each frame where class features are present, 3D CNNs also reveal the frames in which these features are present in greater concentration. 

\begin{figure}[ht]
    \centering
    \animategraphics[loop,autoplay,height=100pt]{8}{tubes_visualizations/heatmaps_2d_-}{01}{16}
    \animategraphics[loop,autoplay,height=100pt]{8}{tubes_visualizations/tubes_3d_-}{01}{16}
\caption{\textbf{Comparison between 2D and 3D saliency.} The main action of the video is \textit{stirring} and it primarily takes place in the middle of the clips. 2D convolutions (left) focus significantly on object appearance without taking into consideration the movements that are performed in the video. This can be seen as every frame in the case of 2D convolutions includes some feature activation. In contrast, 3D convolutions (right) only extract image regions in specific frames where motions are present.}
\label{fig:Fig3}
\vspace{-5mm}
\end{figure}

\section{Conclusions}
\label{sec:conclusions}
In this work, we propose Saliency Tubes as a way to visualize the activation maps of 3D CNNs with relation to a class of interest. Previous work on 2D CNNs establishes visualization methods as a way to increase interpretability of convolutional neural networks and as a supplementary feedback mechanism in terms of dataset overfitting. We build upon this idea for 3D convolutions, using a simple yet effective concept that represents regions in space and time in which the network locates the most discriminative class features.

Additionally, using our visualization scheme, we further validate the notion that 3D convolutions are more effective in learning motion-based features from temporal structures, and they do not only include a larger number of tensor parameters that allow them to achieve better results. We support this by demonstrating how a 2D CNN will focus only on the appearance features per frame for its prediction, whereas a 3D CNN produces a more elaborate spatio-temporal analysis.

\bibliographystyle{IEEEbib}
\bibliography{strings,refs}

\end{document}